\documentclass[sigconf]{acmart}

\usepackage{braket}
\usepackage{qcircuit}
\usepackage[ruled,vlined]{algorithm2e}
\usepackage[english]{babel}
\newtheorem{theorem}{Theorem}

\AtBeginDocument{%
  \providecommand\BibTeX{{%
    \normalfont B\kern-0.5em{\scshape i\kern-0.25em b}\kern-0.8em\TeX}}}

\copyrightyear{2022} \acmYear{2022} \setcopyright{acmlicensed}\acmConference[GECCO '22 Companion]{Geneticand Evolutionary Computation Conference Companion}{July 9--13,2022}{Boston, MA, USA}\acmBooktitle{Genetic and Evolutionary Computation Conference Companion(GECCO '22 Companion), July 9--13, 2022, Boston, MA, USA}\acmPrice{15.00}\acmDOI{10.1145/3520304.3533961}\acmISBN{978-1-4503-9268-6/22/07}

\begin{document}

\title{Quantum Neuron Selection: Finding High Performing Subnetworks With Quantum Algorithms}

\author{Tim Whitaker}
\email{timothy.whitaker@colostate.edu}
\affiliation{%
  \institution{Colorado State University}
  \city{Fort Collins}
  \state{Colorado}
  \country{USA}
  \postcode{80525}
}

\begin{abstract}
Gradient descent methods have long been the de facto standard for training deep neural networks. Millions of training samples are fed into models with billions of parameters, which are slowly updated over hundreds of epochs. Recently, it's been shown that large, randomly initialized neural networks contain subnetworks that perform as well as fully trained models. This insight offers a promising avenue for training future neural networks by simply pruning weights from large, random models. However, this problem is combinatorically hard and classical algorithms are not efficient at finding the best subnetwork. In this paper, we explore how quantum algorithms could be formulated and applied to this neuron selection problem. We introduce several methods for local quantum neuron selection that reduce the entanglement complexity that large scale neuron selection would require, making this problem more tractable for current quantum hardware.
\end{abstract}


\maketitle

\section{Introduction}

Deep neural networks have demonstrated inordinate success across many difficult machine learning tasks in computer vision, natural language processing, and reinforcement learning. Some of the recent advances in these fields have come about in part thanks to larger datasets and larger models. These state of the art models contain billions of parameters and train for thousands of GPU days \cite{brown2020language,zhai2021scaling,dai2021coatnet}. The growing cost of training deep neural networks has led to an increased interest in alternative training methodologies.

Meanwhile, quantum computing continues to garner interest by machine learning researchers as quantum computers have the potential to solve large classes of difficult problems at significantly reduced time complexity compared to classical computers \cite{qiskit}. The mathematical foundations of both machine learning and quantum computing are fundamentally linearly algebraic and there is a lot of hope that quantum methods could enable new ways to drastically improve performance.

Deep neural networks are typically trained with variations of gradient descent. Training data is passed through a network, some loss is determined between the network's output and a target, and the loss is backpropagated through the network according to the gradient with respect to the weights. Modern neural networks repeat this process over millions of training samples for hundreds of epochs.

Recently, there has been some interesting work investigating the subnetworks of these deep neural networks. It's long been known that fully trained deep neural networks can be significantly pruned and maintain high performance after small amounts of tuning \cite{blalock2020state}. However, when those subnetworks are trained from scratch, they fail to achieve the same accuracy they had after dense training and subsequent pruning. A popular line of research titled The Lottery Ticket Hypothesis explored this problem and introduced a method for finding specific subnetworks, in certain network architectures, that train as well as the full size network it is derived from \cite{frankle2019lottery}.

The Lottery Ticket Hypothesis has since spurred many follow up works investigating high performing subnetworks. In particular, some works have introduced methods for finding subnetworks in random models that perform well \textit{without any training} \cite{zhoa2019deconstructing, ramanujan2020hidden}. This approach, which we call \textit{neuron selection}, offers an exciting paradigm for building neural networks in the future by simply pruning weights from large random models.

Discovering the best subnetwork in a large neural network is computationally hard due to exponential combinatoric complexity. Despite the difficulty for finding the \textit{best} subnetwork, these large networks contain so many potential subnetworks that there are likely many that would perform well enough. Current approaches often use heuristics and pseudo-training algorithms to find acceptable subnetworks, where solutions are found by assigning scores to edges and optimizing network graphs with something akin to gradient descent \cite{ramanujan2020hidden, wortsman2019discovering}.

In this paper, we explore applications of quantum algorithms to the neuron selection problem. We demonstrate how neuron selection could be formulated for quantum gate computers and quantum annealing computers and we introduce several local neuron selection algorithms that make these ideas tractable for current quantum systems.

\section{Background}

Training neural networks by slowly adjusting the weights between neurons has long been the de facto standard in deep learning. However, there are several problems that gradient based methods can suffer from, including: insufficient information in the gradients, low signal to noise ratio, architecture bias, and flatness in the activations \cite{shalev2017failures}. These issues make alternative training methods worth investigating for future neural networks.

There also exists strong biological motivation for exploring neural network construction via neuron selection. There is a natural stage of brain development in which a massive amount of neurons, axons, and synapses proliferate. This overabundance creates a competitive environment as these brain cells compete for resources. Many millions of neurons and connections are invariably killed and specialized subnetworks develop as a result. Neural Darwinism, a theory of neuronal group selection, explores how brain function evolves as a result of these selective processes acting on and between groups of neurons \cite{edelman1987neuraldarwinism}.

\subsection{Neuron Selection}

Pruning deep neural networks to find effective subnetworks is an old and established method for optimizing model size and cost \cite{lecun1989obd}. The fields of Neural Architecture Search and Neuroevolution are also closely related as these methods often aim to optimize network architecture. However, applying neuron selection purely as a means for training networks is a relatively new idea in deep learning. 

Weight Agnostic Neural Networks are one such approach to neural architecture search that forgoes the dependence of weight tuning. This work demonstrates how architecture alone could encode solutions to complex problems where networks are constructed with all weights between neurons fixed to a single shared value \cite{gaier2019weight}.

NK Echo State Networks employ neuron selection as a training method by selecting optimal combinations of neurons from a fixed, random, and recurrent reservoir layer \cite{whitley2015nk}. The complex dynamics of the echo state reservoir enables neuron selection to model dynamic time dependent problems without any weight updates. Finding the optimal combinations of neurons is done by connecting a neuron selection layer (probe filter) to an output layer according to an NK-landscape and optimizing this landscape with dynamic programming. 


The Lottery Ticket Hypothesis spawned a large amount of interest in neural network pruning over the last few years. This work introduces an approach to finding extremely sparse subnetworks that train well. First, a dense network is fully trained. Then the network is pruned according to the magnitude of the final weights. The network's weights are then rewound back to the initial values while keeping the sparse structure in tact. The resulting sparse network should then train as well as the dense network it is derived from \cite{frankle2019lottery}.

A followup work, Deconstructing Lottery Tickets, explored the properties of these lottery ticket subnetworks and found that there were subnetworks that were able to perform better than chance without any subsequent weight tuning of the lottery ticket. Zhoa et al. find that when these supermasks are applied to randomly initialized networks, they significantly outperform other random masks and non-masked random networks \cite{zhoa2019deconstructing}.

Recent work has since shown that subnetworks can be found in much larger and more modern neural network architectures than those explored in the original lottery ticket papers. Ramanujan et al. introduce an algorithm to find subnetworks in a randomly weighted WideResNet-50 that matches the performance of a trained ResNet-34 on ImageNet \cite{ramanujan2020hidden}. This work empirically demonstrated that neuron selection could be applied to problems of modern scale and result in extremely accurate networks with absolutely no weight tuning.

Malach et al. laid the theoretical foundations for this line of work in a paper titled, Proving The Lottery Ticket Hypothesis. Assuming some bounded constraints for the norms of weights and inputs, the authors show that a ReLU network of polynomial width and a depth of $2l$ contains a randomly initialized subnetwork that will approximate any trained neural network of width $d$ and depth $l$ \cite{malach2020proving}.




This bound is however much larger than empirically observed in related works. Pensia et al. and Orseau et al. both published proofs that remove the strict assumptions that Malach et al. require and significantly tighten the bounds for the random network to a logarithmic factor \cite{pensia2020subset, orseau2020logarithmic}.

\begin{theorem}
A randomly initialized ReLU network with width $O(d log(dl / min\{\epsilon, \delta\}))$ and depth $2l$, with probability at least $1 - \delta$, can be pruned to approximate any neural network with width $d$ and depth $l$ up to error $\epsilon$ \cite{pensia2020subset}.
\end{theorem}

Several works have also demonstrated that sufficiently overparameterized networks contain many winning lottery ticket subnetworks \cite{malach2020proving, diffenderfer2021multiprize, frankle2019linear}. These results are important as they significantly reduce the computational complexity required to find a single best subnetwork. This is especially important for certain quantum search algorithms, in which multiple solutions can drastically reduce the computation required.

\subsection{Quantum Optimization}

There is a lot of interest and excitement in the application of quantum algorithms to machine learning, as quantum computers have the potential to solve extremely large and difficult problems that would be intractable on classical computers.

Quantum Annealing and Quantum Gate Computing are the two primary paradigms under which quantum computers operate \cite{gyongyosi2019survey}. Quantum annealers are specialized pieces of hardware that are tailored to specifically formulated optimization problems. Quantum gate computers are more generalized and are analogous to classical gate circuits, yet they make use of special quantum gates that act on qubits.

Below we introduce some of the most popular methods for quantum optimization on both quantum annealers and quantum gate computers.

\subsubsection{Grover's Search}

Grover's algorithm is a prominent quantum algorithm that uses amplitude amplification to perform unstructured search. Consider searching through a list of $N$ items. In order to find a specific, unique item, one would need to check on average $N/2$ locations on a classical computer. In the worst case, one would need to check all $N$ locations. On a quantum computer, Grover's algorithm can find the marked item in approximately $\sqrt{N}$ iterations. Grover's algorithm offers quadratic runtime speedup over naive classical methods and can save significant computation with large search spaces \cite{grover1996fast}.

The generality of this approach lies in the fact that it is often difficult to find the correct solution, but easy to verify a solution. Grover's algorithm begins by placing all qubits into a uniform superposition $\ket{s}$. An oracle is then queried which flips the amplitude of "marked" states. This is implemented with a unitary operator that returns -1 for any marked state $\omega$ and 1 otherwise.
\[
U_{\omega} \ket{x} = 
\begin{cases}
    \phantom{-}\ket{x} & if \ x \neq \omega \\
    - \ket{x} & if \ x = \omega
\end{cases}
\]

The result of $U_{\omega}$ is then passed through a diffusion operator $U_d$ which flips all amplitudes in $\ket{s}$ about the mean.
\[
U_d = 2 \ket{s} \bra{s} - I
\]

The oracle and diffusion operations are then repeated several times to amplify the probability that the marked state(s) will be measured. After $t$ iterations, the state of the system can be described as
\[
\ket{\psi_t} = (U_f U_d)^t \ket{s}
\]

\subsubsection{Quantum Annealing}

Many combinatorial optimization problems in quantum computing are solved with methods based on the Adiabatic Theorem. These approaches slowly evolve a quantum system over time from some initial quantum state towards a corresponding ground state that encodes the solution to the problem. Evolution of the system is primarily governed by a Hamiltonian operator, which is an operator associated with the energy state of a system.

These methods are closely related to Ising models, which are used to study magnetic dipole moments of atomic spins \cite{brush1967ising}. In these models, discrete variables are organized into a lattice and described as spins which can be in either an up or down state ($+1$ or $-1$). The local structure of Ising models assumes that spin states interact only with their neighbors. This formulation makes adiabatic methods very popular for satisfiability problems where the interactions of local neighborhood structures can be exploited.

Quantum Annealing is one of the most prominent adiabatic optimization method where the evolution of a system can be described as \cite{defalco1988annealing, apollini1989combinatorial}:
\[
H(t) = (1 - \frac{t}{T}) H_0 + \frac{t}{T} H_C
\]

\noindent where $H_0$ is a problem independent mixing Hamiltonian that applies weak perturbations to the system in order to aid in exploration, and $H_C$ is the problem (or cost) Hamiltonian that describes the energy of the system. Evolving the system over a long time period $T$ results in convergence to the ground state of $H_C$, which encodes the solution to the problem. The problem Hamiltonian $H_C$ then can be constructed by evaluating a given cost function $C(x)$:
\[
H \ket{x} = C(x) \ket{x}
\]

\subsubsection{Variational Circuits}

These are hybrid approaches that combine quantum states with classical optimization. In these algorithms, a quantum system is parameterized by some set of variables $\theta$. The quantum system is measured, some loss is calculated and the parameters are updated with classical algorithms. This hybrid approach enables applications of quantum systems to wide varieties of problems.

Variational Quantum Eigensolver (VQE) and Quantum Approximate Optimization Algorithm (QAOA) are perhaps the two most popular variational methods \cite{peruzzo2014variational, farhi2014qaoa}. These approaches are closely related to the adiabatic methods based on quantum annealing defined above, however these methods operate with quantum logic gates. As in quantum annealing, both methods rely on constructing a problem Hamiltonian $H$ that describes the energy state of a system.

VQE works by first preparing an initial state according to some set of parameters $\ket{\psi(\theta)}$, often called an ansatz. The expectation of the quantum state is measured $\bra{\psi(\theta)} H_C \ket{\psi(\theta)}$ and the parameters $\theta$ are then updated using a classical algorithm.

In QAOA, a special set of unitary operators are used to alter the ansatz. First, a quantum state $\ket{\psi(\gamma, \beta)}$ is prepared using unitary operators of the form $U(\beta) = e^{-i \beta H_0}$ and $U(\gamma) = e^{-i \gamma H_C}$ with the problem Hamiltonian $H_C$ and a mixing Hamiltonian $H_0$. These unitaries are applied repeatedly in blocks $p$ times to some initial state $\ket{\psi_0}$.
\[
\ket{\psi(\beta, \gamma)} = U_0(\beta)U_0(\gamma) ...  U_p(\beta)U_p(\gamma)\ket{\psi_0}
\]

Then, an expectation is computed using the given parameters $\gamma, \beta$ with respect to the problem Hamiltonian $H_C$.
\[
F_p(\gamma, \beta) = \bra{\psi_p(\gamma, \beta)} H_C \ket{\psi_p(\gamma, \beta)}
\]

Finally, a classical optimizer is used to optimize the parameters $(\gamma, \beta)$. This process is repeated until some threshold or convergence criteria is met. In general, variational and annealing algorithms do not have computational complexity guarantees over classical methods \cite{aaronson2018introduction}.

\section{Quantum Neuron Selection}

\begin{figure*}
    \centering
    \includegraphics[width=\textwidth]{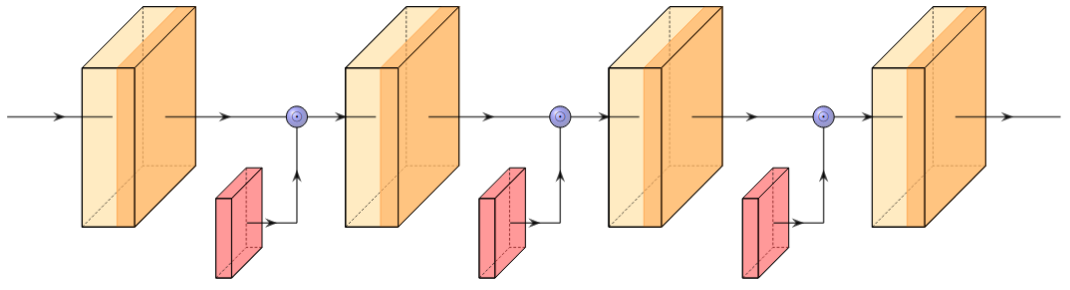}
    \caption{An illustration of a hybrid quantum-classical neural network for use with the quantum edge-popup algorithm. Quantum circuits (red) are evaluated and the resulting mask is applied to the output of each layer (yellow) using an elementwise Hadamard product ($\odot$). The weights of the layers are fixed, and only the rotation parameters $\theta$ for each quantum circuit are optimized.}
    \label{fig:my_label}
\end{figure*}

The application of quantum algorithms to neuron selection offers a lot of potential opportunities for exploring new methods of training machine learning models. We hope that this work introduces interesting ideas for which new training methods and network architectures could develop.

\subsection{Preliminaries}

Consider a feed forward neural network $F: \mathbb{R}^n \rightarrow \mathbb{R}$ consisting of $d$ layers $\{f_1, ... , f_d\}$ parameterized with weights $\{W_1, ... , W_d\} \in \mathbb{R}$, masks $\{M_1, ..., M_d\} \in \mathbb{B}$, and biases $\{b_1, ..., b_d\} \in \mathbb{R}$.
\begin{align*}
F(x) &= f_d \circ ... \circ f_1(x) \\
f_i(x) &= \sigma(x \cdot (M_i \odot W_i) + b_i) \\
f_d(x) &= x \cdot (M_d \odot W_d) + b_d
\end{align*}

\noindent where $W_i \in \mathbb{R}^{MXN}$ is a fixed weight matrix for layer $i$ that connects $M$ neurons in the previous layer to $N$ neurons in the current layer. $M_i \in \{0, 1\}^{MXN}$ represents the subnetwork mask that is applied to the weights with a Hadamard product $\odot$. This masked weight matrix is multiplied with the layer input $x \in \mathbb{R}^M$ and the result is then summed with a bias vector $b_i \in \mathbb{R}^N$ and passed to a nonlinear activation function $\sigma$.

Evaluation of a given network can be done with a variety of loss functions that vary according to the task at hand. In this case, consider a regression task in which the goal is to predict a single real value $y \in \mathbb{R}$. The loss for a given input sample $x$ can then be described with the L2 distance between the network's prediction $F(x)$ and the true value $y$.
\[
\mathcal{L}(x, y) = ||F(x) - y||_2
\]

The optimization objective then is to minimize this loss over a training dataset consisting of $N$ samples by optimizing the binary bitmask $M$.
\[
\underset{M}{\text{minimize}} \ \mathcal{L} = \frac{1}{N} \sum_{i=1}^N ||F(x_i; M) - y_i||_2
\]

\subsection{Global Optimization}

Quantum Neuron Selection can be naturally framed as an unstructured search or combinatorial optimization problem. For each parameter, a qubit is used to determine whether that given parameter should be kept or not. Representing subnetwork masks with a string of qubits allows this formulation to be trivially applied to any of the of popular quantum algorithms described in the background section, including Grover's Search, Quantum Annealing, Quantum Approximate Optimization, and Variational Quantum Eigensolver. However, there are some challenges in applying quantum optimization methods to problems at the scale of modern neural networks. 

The crux of these problems primarily lie in the construction of an efficient oracle or target Hamiltonian. While theoretically simple using the loss and Hamiltonian descriptions in the previous section, this is actually quite difficult in practice. The oracle needs to implemented as a quantum circuit that can store data, perform matrix multiplication, and apply activation functions on floats which consist of many bits per value. Any classical computation can be compiled down into reversible boolean logic gates, however, the current state of software compilation makes this practically difficult \cite{qiskit}.

\begin{algorithm}[!h]
\SetAlgoLined
 \KwIn{W, set of fixed weights}
 \KwIn{M, mask representing the active subnetwork}
 \KwIn{V, validation dataset of length N}
 \KwIn{$\epsilon$, loss threshold}
 $F \leftarrow initializeNetwork(W, M)$ \\ 
 $\mathcal{L} \leftarrow 0$ \\
 \For{(x, y) in V}{
    $\mathcal{L} \leftarrow \mathcal{L} + \frac{1}{N}||F(x) - y||_2$ \\ 
 }
 \KwOut{$\mathcal{L} < \epsilon$}
 \caption{Neuron Selection Oracle}
\end{algorithm}

Additionally, constructing a target Hamiltonian can be extremely costly and memory intensive. The size of the Hamiltonian matrix is $2n x 2n$ for $n$ parameters and implementing the Hamiltonian for any moderately sized network would take far too long. It is also necessary to translate subnetwork evaluation and loss to a quadratic unconstrained binary optimization (QUBO) form, which implements problem constraints as penalties in order to aid in optimization. Sasdelli et al. demonstrate how this can be done with small, binary neural networks \cite{sasdelli2021annealing}.

Quantum computers are fragile and limited to small numbers of qubits. The largest quantum gate computer at the time of this writing is the IBM Eagle processor which consists of 127 qubits with a a 64 bit quantum volume \cite{collins2021ibm}. Quantum annealers are able to operate with far more qubits, with the current state of the art being the D-Wave Advantage, which consists of 5,640 qubits \cite{mcgeoch2022advatage}. However, quantum annealers do not have the same computational complexity or generality of quantum gate computers and it's unclear whether they can offer the same types of computational advantages over specialized classical algorithms.

Placing large amounts of qubits into superposition will require significant advances in quantum hardware and software. For these reasons, we introduce several ideas for implementing local neuron selection that may make this problem more tractable on near term quantum systems.

\subsection{Quantum Edge-Popup}

Edge-popup is a classical algorithm for finding good subnetworks by using gradient heuristics \cite{ramanujan2020hidden}. Parameters are assigned a score, and scores are updated for each training sample while the weights stay fixed to their random initial values. The subnetwork is chosen by selecting the top k\% of the parameters according to their scores within their layers.

We explore how we can adapt edge-popup to a quantum framework by implementing a hybrid variational circuit that can be plugged in to classical neural network layers. Rather than assigning a score to each parameter, we instead represent the score with a quantum state that governs how likely the parameter is to be kept.

We begin by constructing a parameterized quantum circuit for each layer in our network. When measured, the circuit will collapse to the subnetwork mask $M$. The circuit consists of $N$ qubits for each weight in layer $i$. Each qubit is placed into a uniform superposition which is then manipulated with rotation gates according to learned parameters $\theta$. We bound the values for $\theta$ to the range $[- \pi / 2, \pi / 2]$.
\[
\Qcircuit @C=1.2em @R=0.8em {
& \lstick{\ket{0}} & \gate{H} & \gate{R_y(\theta_0)} & \meter & \qw \\
& \lstick{\ket{0}} & \gate{H} & \gate{R_y(\theta_1)} & \meter & \qw \\
\vdots \\
\\
& \lstick{\ket{0}} & \gate{H} & \gate{R_y(\theta_n)} &  \meter & \qw}
\]

A forward pass for a given training sample $x$ is performed and the loss $\mathcal{L}$ is determined between the networks output $F(X)$ and the true label $y$. The gradient is then calculated for the loss with respect to the quantum circuit parameters $\theta$. Since the gradient for much of the network is 0 due to the subnetwork mask, we use a straight through estimator in which the backward pass assumes all connections were used \cite{ramanujan2020hidden}. This approach ensures that the gradient does not vanish for early layers in the network. The parameter update equation can be described as:

\[
\theta_{uv} = \theta_{uv} - \alpha \frac{\partial \mathcal{L}}{\partial \mathcal{I}_{v}} Z_u w_{uv}
\]

\noindent where $\theta_{uv}$ is the rotation parameter for the edge that connects neuron $u$ in the previous layer to neuron $v$ in the current layer. $\alpha$ is the learning rate. $\mathcal{I}_v$ is the input of neuron $v$. and $Z_U w_{uv}$ is the weighted output of neuron $u$.





\subsection{Layerwise Knowledge Distillation}

\begin{figure}
    \centering
    \includegraphics[width=\columnwidth]{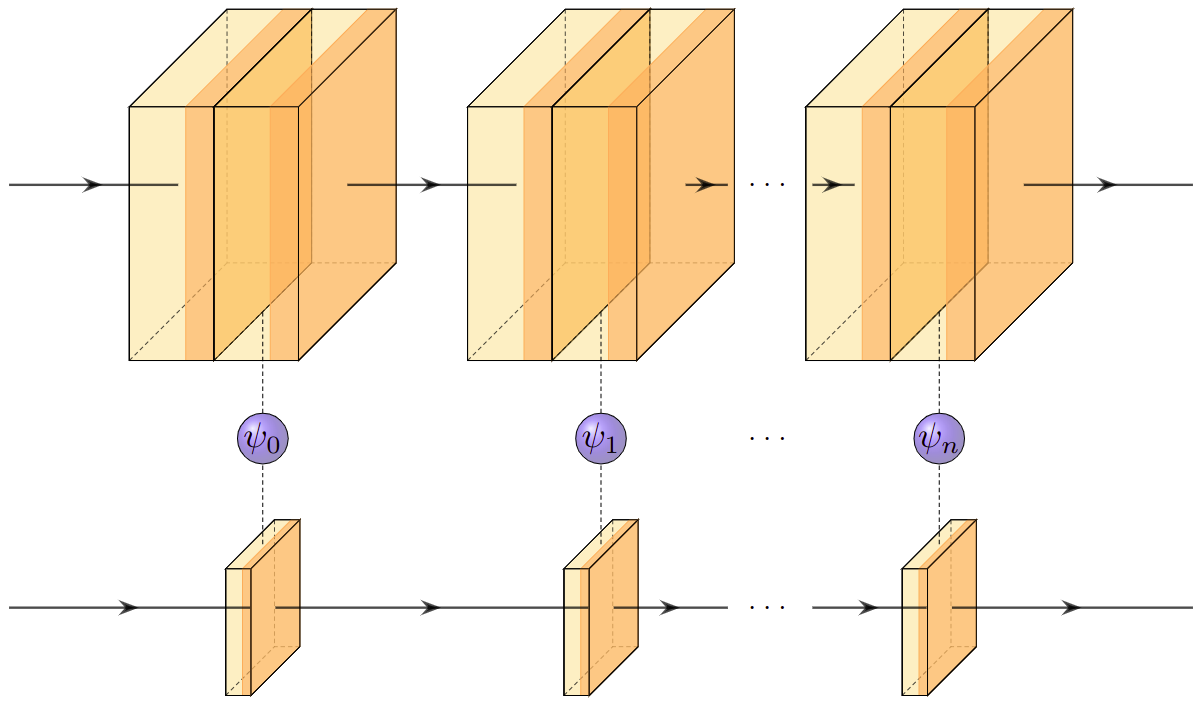}
    \caption{An illustration of Layerwise Knowledge Distillation. A random student network (top) is successively optimized with a quantum neuron selection algorithm (middle) according to the layerwise representations of a trained teacher network (bottom).}
    \label{fig:my_label}
\end{figure}

Knowledge distillation is a prominent algorithm for supervised learning problems that is able to transfer knowledge learned from a large model to a significantly smaller model \cite{hinton2015distilling}. Rather than training a network on the dataset itself, knowledge distillation trains a network on the outputs of another fully trained network. This approach works on the idea that a small network may not be able to capture complexities in the raw data itself, but it can learn from the decisions that a trained larger network is able to make.

This approach can be used to implement neuron selection at a layerwise level. Rather than finding subnetworks in a large random model according to a validation dataset, we can instead use a trained \textit{teacher} network as a guide. This way, we can find subnetworks layer by layer, by optimizing the random \textit{student} network to imitate the layer representations of the trained teacher network.

Consider two neural networks, a small fully trained teacher $F_t$ and a large randomly initialized student network $F_s$. Assuming the student network is sufficiently wide and twice as deep as the target network, according to the lottery ticket proofs described in the background section above, it is reductively possible that every two layers in the student network should contain a subnetwork that can approximate a single corresponding layer in the fully trained teacher network.

\begin{algorithm}[!h]
\SetAlgoLined
 \KwIn{$W_S$, $W_T$, random student weights and trained teacher weights}
 \KwIn{$V$, validation dataset}
 $F_S \leftarrow initializeNetwork(W_S)$ \\
 $F_T \leftarrow initializeNetwork(W_T)$ \\ 
 $nlayers \leftarrow size(F_T)$ \\
 $M \leftarrow []$ \\
 \For{i in 1 to nlayers}{
   $y_T \leftarrow getActivations(F_{T_{i}}, V)$ \\
   $y_S \leftarrow getActivations(F_{S_{i:i+1}}, V)$ \\
   $M_i \leftarrow quantumOptimization(y_T, y_S)$ \\
 }
 \KwOut{$M$, optimized bit mask for the student network}
 \caption{Simplified Knowledge Distillation}
\end{algorithm}

We start by running a validation set through the teacher network, where the outputs at each layer are recorded for all the training samples. We then perform quantum neuron selection on the student network $F_s$ two layers at a time, where the goal is to find the subnetwork for those two layers that approximates the target layer in the teacher network $F_t$. We can use several of the quantum optimization methods described in the previous sections to find the optimal bit string that maximizes similarity between the teacher network's layer output and the student network's layer output.



\subsection{Reducing Entanglement Complexity With NK Landscapes}

Along with applying neuron selection to localized network structures, it's possible to impose constraints on neural network connectivity in order to reduce the entanglement complexity of quantum systems. NK Echo State Networks provide an elegant framework for implementing localized neuron selection with a quantum classification layer.

Echo State Networks belong to a class of recurrent networks called reservoir computing models. In the simplest case, an echo state network consists of an input layer, a reservoir, and an output layer. During training, only the weights associated with the output layer are updated \cite{jaeger2004echo}.

The reservoir contains a moderate amount of sparse and randomly connected neurons. At each iteration, the output of the reservoir is fed back in to the reservoir in a recurrent fashion. The connectivity of the neurons and the weights associated are initialized such that the spectral radius of the weight matrix is less than 1. This property ensures that the recurrent state that is passed back into the reservoir slowly decays over time, like an echo.

Echo State Networks can be succinctly described with the following system equations:
\begin{align*}
z_{res}(x_t) &= W_{res} z_{res}(x_{t-1}) + W_{in} x \\
y(x_t) &= \sigma(W_{out} z_{res}(x_t))
\end{align*}

\noindent where $z_{res}(x_t)$ is the reservoir state at iteration $t$ for a given input $x$. The input, reservoir, and output weights are $W_{in}$, $W_{res}$, and $W_{out}$ respectively. The output, $y(x_t)$, is gotten by multiplying the reservoir state with the output weights and applying a nonlinear activation function $\sigma$.

Echo State Networks are fast, simple to implement and well suited for dynamic time series data, thanks to the chaotic and cyclical behavior of the reservoir.

NK Echo State Networks are a variation on Echo State Networks that utilize neuron selection by incorporating a probe filter layer between the output layer and the reservoir \cite{whitley2015nk}. The probe filter layer contains $N$ neurons that each connect randomly to neurons within the reservoir. Turning off and on different combinations of neurons in the probe filter layer results in extremely diverse reservoir dynamics as the neural circuitry is altered.

The network is trained only by optimizing a bitmask $M$ which describes which probe filter neurons are turned on or off. The system dynamics are then described as:
\begin{align*}
z_{res}(x_t) &= W_{res} z_{res}(x_{t-1}) + W_{in} x \\
z_{pf}(x_t) &= M \circ W_{pf} z_{res}(x_t) \\
y(x_t) &= \sigma(W_{out} z_{pf}(x_t))
\end{align*}

\noindent where $z_{pf}$ is the output of the probe filter layer, $W_{pf}$ are the weights of the probe filter layer, $M \in \{0,1\}^n$ is an n-dimensional bit vector that describes which neurons to turn on and off, and $\circ$ is the Hadamard product. 

The probe filter layer is connected to the output layer according to an NK-landscape, where $N$ neurons in the probe filter layer are connected to exactly $K$ neurons in the output layer. This formulation allows each separate output neuron to act as a k-bounded pseudo-boolean subfunction $f_i(x)$ in which the search space is restricted to the size of $K$.

For each output neuron, performance is recorded for every possible on/off pattern for the $K$ neurons in the probe filter layer that connect to it. This results in a look-up table for performance evaluations with $2^K N$ entries. The bitstring is then optimized using dynamic programming in order to minimize the loss when all outputs are averaged together.
\begin{align*}
F(x) &= \frac{1}{N} \sum_{i=1}^N y_i(x) \\
\mathcal{L}(x, y) &= ||F(x) - y||_2
\end{align*}

\begin{figure}
    \centering
    \includegraphics[width=\columnwidth]{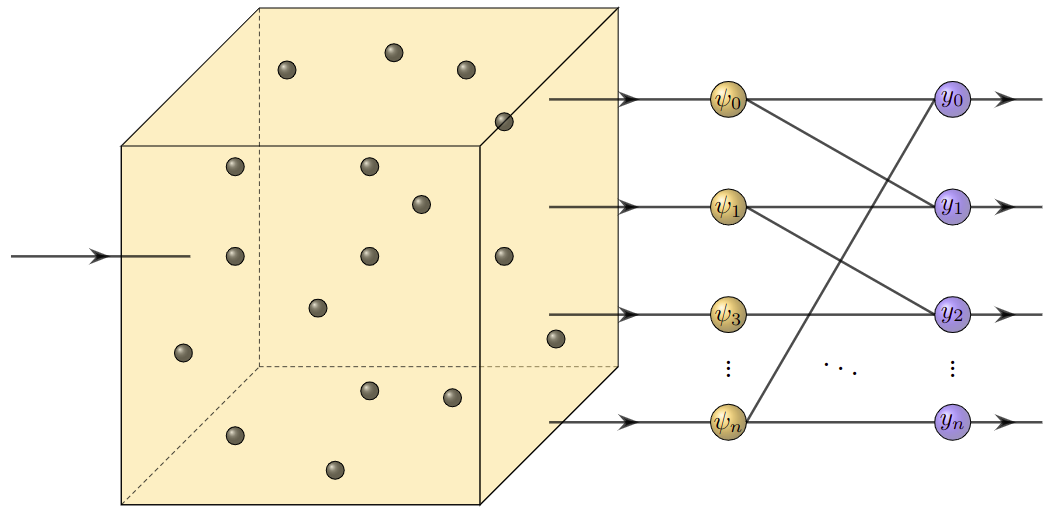}
    \caption{A network architecture diagram of an NK Echo State Network. A recurrent reservoir is connected to a probe filter layer which is then connected to an ensemble of output nodes $y$. The probe filter layer $\psi$ is connected to the output layer according to an NK landscape and is able to be optimized with an efficient quantum neuron selection algorithm.}
    \label{fig:my_label}
\end{figure}





NK Echo State Networks make use of clever ensembling scheme in which output neurons are duplicated and connected to different subsets of neurons in the probe filter layer. This approach allows for a significant reduction in entanglement complexity, as only combinations of the $K$ neurons that connect to an output neuron needs to be exhausted.

This connective structure allows for a natural transition to quantum optimization. Rather than generating a lookup table and optimizing with dynamic programming, we can now apply any of the previously described methods to each output neuron individually, where the total number of qubits for each circuit is only $K$. Applying Grover's Search then can reduce the runtime complexity from $\mathcal{O}(2^K N)$ to $\mathcal{O}(\sqrt{2^k} N)$.

Neuron selection works particularly well for echo state networks, due to the complex dynamics of the recurrent reservoir. However, this approach and connective structure is quite generic, and a quantum probe filter layer and ensemble output could be embedded in network architectures of any type. NK landscapes enable computationally efficient pseudo-ensembles where layers of any width can be optimized with quantum neuron selection.

\vspace{0.35in}

\section{Conclusions}

Several works have demonstrated that large neural networks contain subnetworks that perform as well as fully trained networks \cite{ramanujan2020hidden, malach2020proving, pensia2020subset, orseau2020logarithmic}. However, there is no efficient classical algorithm for finding the best subnetwork in a given network. We discuss several prominent quantum optimization algorithms, including Grover's Search, Quantum Annealing, Variational Quantum Eigensolver, and Quantum Approximate Optimization, and we explore how quantum neuron selection could be formulated to be solved with these algorithms.

Due to the scale of deep neural networks and the fragility of current quantum systems, pure quantum neuron selection will require significant advances in both hardware and software. Nevertheless, we introduce several approaches for performing local quantum neuron selection that are tractable for current quantum systems. The first is Quantum Edge-Popup, a variational method that embeds parameterized quantum circuits into classical neural networks. The second approach is based on Layerwise Knowledge Distillation, which finds subnetworks in a random student subnetwork by learning to approximate the representations of a smaller trained teacher network. Lastly, we offer an exploration of NK Echo State Networks, where we discuss how K-bounded connectivity patterns can significantly reduce the entanglement complexity required for quantum systems. This approach enables quantum neuron selection to be applied to probe filter layers of arbitrary width.

Quantum Neuron Selection is an exciting new paradigm for constructing deep neural networks. There is strong biological motivation for exploring neuron selection as a training mechanism \cite{edelman1987neuraldarwinism} and we think this approach could potentially enable new methods for building and designing neural networks of the future.

\bibliographystyle{ACM-Reference-Format}
\bibliography{bibliography}

\end{document}